\def\year{2021}\relax
\documentclass[letterpaper]{article} 
\usepackage{aaai21} 
\usepackage{times} 
\usepackage{helvet} 
\usepackage{courier} 
\usepackage[hyphens]{url} 
\usepackage{graphicx} 
\usepackage{graphicx} 
\usepackage{natbib} 
\usepackage{caption} 
\usepackage[switch]{lineno}  %
\usepackage{amsfonts}
\usepackage{amsmath}
\usepackage[export]{adjustbox}
\usepackage{booktabs}
\usepackage{color}
\usepackage{float}

\newcommand{\todo}[1]{}
\renewcommand{\todo}[1]{{\color{red} TODO: {#1}}}

\title{Uncertainty Estimation for Community Standards Violation in Online Social Networks}

\author{\normalfont{Narjes Torabi} ,
   {Nimar S. Arora} ,
   {Emma Yu} ,
   {Kinjal Shah} ,
   {Wenshun Liu} ,
   {Michael Tingley}\\
   Facebook Inc, USA
   }
\begin{document}
\maketitle

\begin{abstract}
Online Social Networks (OSNs) provide a platform for users to share their thoughts and opinions with their community of friends or to the general public.
In order to keep the platform safe for all users, as well as to keep it compliant with local laws, OSNs typically create a set of community standards organized into policy groups, and use Machine Learning (ML) models to identify and remove content that violates any of the policies.
However, out of the billions of content that is uploaded on a daily basis only a small fraction is so unambiguously violating that it can be removed by the automated models.
Prevalence estimation is the task of estimating the fraction of violating content in the residual items by sending a small sample of these items to human labelers to get ground truth labels.
This task is exceedingly hard because even though we can easily get the ML scores or features for all of the billions of items we can only get ground truth labels on a few thousands of these items due to practical considerations.
Indeed the prevalence can be so low that even after a judicious choice of items to be labeled there can be many days in which not even a single item is labeled violating.
A pragmatic choice for such low prevalence, $10^{-4}$ to $10^{-5}$, regimes is to report the upper bound, or $97.5\%$ confidence interval, prevalence (UBP) that takes the uncertainties of the sampling and labeling processes into account and gives a smoothed estimate.
In this work we present two novel techniques Bucketed-Beta-Binomial and a Bucketed-Gaussian Process for this UBP task and demonstrate on real and simulated data that it has much better coverage than the commonly used bootstrapping technique.
Our work has been deployed in a large scale OSN and it provides daily UBP estimates on hundreds of policy groups.
These estimates are used by the OSN to measure the impact of its internal counter-measures as well for reporting to government regulatory bodies.

\end{abstract}

\section{Introduction}
\label{sec:introduction}
Online Social Networks (OSN) \cite{garton1997studying} fill an important human need to freely communicate ideas and opinions.
Many OSNs such as LinkedIn, Facebook and Twitter have been able to reach a global scale and provide us unprecedented access to a broad and relevant audience.
These networks can be used to express mundane ideas or to organize grass roots political movements \cite{bruns2013arab} with long lasting repercussions.
At present, the role of these networks is still evolving as their reach continues to increase in various spheres such as politics, medicine, entertainment, and journalism, in addition to inter-personal communication.

As a consequence of their widespread reach, OSNs have attracted a broad spectrum of malicious actors wishing to pervert the public discourse.
These actors attempt to use the networks to spread false stories, suppress voter turnout, incite violate, express hatred towards minorities, sexually exploit minors, or financially defraud susceptible segments of the population among a wide variety of unacceptable behavior.
Although human societies have been dealing with similar negative actions and behaviors for millennia including continually redefining what constitutes a crime \cite{hanawalt1999medieval}, the broad reach of OSNs has given rise to the need for a new definition of acceptable online behavior.
These acceptable actions are typically called \emph{community standards} \cite{Youtubestandard, FBstandard} and some version of these have been formally adopted by almost every OSN.
The standards are divided into \emph{policy groups} each of which is targeted towards a certain class of violation such as pornography, hate speech, fake accounts, etc.
Depending on the severity of the violation, the OSN may choose to simply block the offending content or to refer the user to local law enforcement.

The sheer scale of the task of policing billions of new content items every day necessitates the use of automated Machine Learning (ML) models to detect and remove violations.
In a typical setup, a separate ML classifier is trained for each policy group.
This classifier outputs a score between $0$ and $1$ which is to be interpreted as the probability that a given content violates a policy group.
However, these ML models are not perfect.
The training of ML classifiers requires vast amounts of high quality labeled data, but in practice it is only feasible to obtain roughly a thousand new labels for each of the hundreds of policy groups each day.
Since the daily volume of data is too small for reliably training a classifier, the data from a trailing time window is  typically aggregated together to train a new daily classifier.
In effect, the classifier is providing a trailing prediction, which may miss emerging trends.

Additionally, the data itself can suffer from extreme class imbalances with overall violation probabilities being as low as $10^{-5}$ for certain policy groups.
In the presence of under-represented data and severe class distribution skews, standard ML algorithms tend to be overwhelmed by the majority class and ignore the minority class. This under-performance is a consequence of classifiers seeking to achieve good performance over a full range of instances by maximizing accuracy over the entire population \cite{guo2008class}.
Optimizing quantities such as error rate does not take the data
distribution into consideration and is not useful for unbalanced datasets.
As investigated in \citep{Japkowicz02theclass, 5128907}, the degree of imbalance is not the only factor that hinders
learning and dataset complexity also plays a role in determining classification deterioration. \citet{Sofiaarticle} categorized class distribution and non-equal misclassification
costs (when misclassification comes with high penalty) as characteristics that affect performance of traditional ML.

The ML classifier scores are used in two ways.
The most obvious violating content, i.e. those with high scores, are typically flagged for human review or automatically removed.
For the remaining/residual items, the classifier scores are used to estimate the \emph{prevalence} of the policy violation, i.e. the fraction of contents that violate the policy.
These prevalence numbers are reported to local authorities in compliance with regulatory laws.
These numbers may also be used internally by the OSNs to validate the effectiveness of counter-measures to limit the spread of policy violations.

For the purposes of removing obviously violating content, uncalibrated scores are sufficient.
However, these scores are unsuitable for directly estimating prevalence on the residual items.
Even small systematic errors can get amplified when extrapolated over billions of items.
In this paper, we investigate the problem of estimating the true prevalence by calibrating the ML classifiers using the most recent subset of the labeled data.
For low prevalence policies, the recent labeled data may not have even a single violating example.
This sparsity of positive labels makes calibration and estimation an extremely challenging problem.

In this paper, we explore statistical techniques to get the confidence intervals on the prevalence, and in particular the $97.5$ percentile prediction, or the \emph{Upper Bound Prevalence} (UBP).
For low prevalence policies, the UBP is the number that is actually reported to the regulatory agencies, and this is our primary focus here.
The main contribution of this paper are two non-parametric Bayesian approaches that can give a much more accurate estimate of UBP than existing techniques.
Further our methods are designed to be practical to deploy and indeed they are currently deployed in one of the largest OSNs.
In fact, one of the proposed inference techniques can be implemented entirely within a SQL query.

We describe the related research in Section~\ref{sec:related}, and in Section~\ref{sec:approach} we formally define the problem and our proposed approach. Finally, in Section~\ref{sec:results} we show the results of experiments on actual OSN data for two policies.

\section{Related Work}
\label{sec:related}
A lot of research effort has been focused on studying the problem of imbalanced class predictions.
A prominent class of techniques is the so called \emph{up/down sampling}~\cite{seiffert2009rusboost}, which is designed to handle the disproportionate distribution of binary classes in imbalanced datasets.
Despite showing promising results, up-sampling introduces statistics linked
to non-existing labels and down-sampling prevents us from viewing certain characteristics of real labels. Similar to \citet{Imbalanced_target_prediction}, in this study we
aim to utilize labeled data to train a model and provide robust UBP estimation inferred from the \emph{original} imbalanced prevalence.
All these studies indicate that the imbalanced learning problem
requires developing new understandings, principles, algorithms and tools to more efficiently transform the vast amounts of un-labeled data into information and knowledge representation that is sensitive to rare classes.

Other research has focused on developing bucketing based methods aiming to get the output of a classifier and transforming it to a well calibrated class membership probability.
In the histogram binning approach ~\cite{histogrambinning}, labeled items are sorted by classifier scores and bucketed into equal size bins.
The violating probability of an unlabeled item is predicted to be the fraction of violating items in the bucket corresponding to the unlabeled item's score.
In our low prevalence setting most of these buckets will have zero violating items and hence this will lead to a gross underestimation of the prevalence.

Isotonic regression ~\cite{isotonic} is another non-parametric form of regression which assumes
that the mapping function between classifier's uncalibrated score and the true violation probability is a non-decreasing (or in some versions a strictly increasing) function.
This approach can be viewed as a binning algorithm where the position of
boundaries and the size of bins are selected according to how well the classifier ranks examples in the training data.
Unfortunately, the monotonic assumption of the mapping function is not a practical assumption in an OSN where hundreds of ML classifiers are trained (one per policy) by independent teams using a variety of techniques.
Also, it is nontrivial to estimate the parameters of this form of regression.

Finally, bootstrapping~\cite{Raschka2018ModelEM} is a fairly general frequentist technique for uncertainty estimation that is generalizable in a variety of settings.
In this approach we can bypass any attempt to calibrate the classifier scores and directly estimate the prevalence by drawing bootstrap samples from the labeled data.
Similar to other methods discussed above, this also fails in the low prevalence regime when the labeled data may not be a perfect representation of the desired data distribution.
For example if the labeled dataset has no violating example, which has non-zero probability of occurring on any given day, this technique produces an underestimate.

\section{Proposed Approach}
\label{sec:approach}
\subsection{Problem Statement}
\label{sec:statement}
Every piece of content on an OSN whether it be a status update, a video, a conversation, or even a user profile is processed by hundreds of ML classifiers (one per policy group) to see if it violates any of the community standards.
Let's assume that one such classifier is presented with $M$ content items, $1\ldots M$, and it assigns each item a continuous-valued score $x_i$ in range $[0, 1]$.

Further, we assume that the binary label $y_i \in \{0, 1\}$ indicates whether the content $i$ violates the policy corresponding to this classifier, and that we are given a subset $S$ of the items of size $N$ for which the true label $y_i$ is known for  $i \in S$.
In practice, the labeled subset is relatively small, i.e. $M \gg N$.
Our task is to estimate the prevalence defined as,
\begin{align}
\pi &= \frac{1}{M} \sum_{i=1}^M y_i \label{eqn:prevalence}
\intertext{
    Further, Upper Bound Prevalence is defined as,
}
u_{\pi} &= \min \{u \ | \ P(\pi <= u) \ge .975\}
\intertext{
    Where, we consider the posterior probability of $\pi$ over a probabilistic model that captures the errors in the classifier scores.
    In this context, we define the calibration curve as,
}
f(s) &= P (y_i=1 | x_i=s)
\end{align}
Our inference task is to estimate $\pi$, $u_{\pi}$, and also optionally $f(\cdot)$ given $\mathbf{x}$, $S$, and $y_S$.
Note that the task of choosing an optimal sample $S$ is beyond the scope of this work.

\subsection{Bayesian perspective and bucketing}
\label{sec:bayesian}
The Bayesian approach \cite{howson2006scientific} is a very natural fit for situations with relatively small amounts of data.
The centerpiece of this approach is a generative probabilistic model (also known as the prior) which is often supplied by a domain expert.
Informally, the prior helps ``fill in" the gaps in the region where we don't have enough data.
This knowledge-based modeling approach helps supplement a small amount of data, and leads to robust estimates even in the face of uncertain, incomplete or imbalanced data.

For the prevalence estimation problem, we make the a-priori assumption that the calibration curve, $f(\cdot)$, is a smooth function.
Our general approach will be to compute the posterior distribution of this calibration curve, and use it to estimate the prevalence and UBP.
In this regard, we need a representation of this curve that is expressive enough to capture its fluctuations without being so verbose as to make the approach computationally infeasible.
We have adopted here the pragmatic approach of subdividing the score range into $K$ equal width buckets and treating the calibration curve as constant inside a bucket.

We use the following pre-processing step of bucketing the input data that is used in all of our proposed approaches.
\begin{enumerate}
  \item We will create $K$ buckets $B_1,\ldots,B_k$ such that $B_k$ represents scores in the range $[\frac{k-1}{K}, \frac{k}{K})$.
  \item Define $I_k = \{\ i \ |\ x_i \in B_k\}$ and calculate $w_k = \frac{|I_k|}{M}$, which is the weight of the bucket $k$.
  \item Define $S_k = I_k \cap S$, $n_k = |S_k|$, and calculate $n_k^{+} = \sum_{i \in S_k} y_i$ and $n_k^{-} = n_k - n_k^{+}$.
  \item Define $p_k$ as the random variable that represents the assumed constant value of $f(\cdot)$ in $B_k$.
\end{enumerate}

\subsection{Bucketed-Beta-Binomial}\label{sec:BBB}
Bucketed-Beta-Binomial (BBB) model assumes that the prevalence in each bucket, $p_k$, is drawn from a Beta distribution and the number of violating items in each bucket is drawn from a Binomial distribution conditional on $p_k$.
Given hyperparameters $a, b > 0$, we define the model as,
\begin{align}
  p_k &\sim Beta(a, b) & k = 1\ldots K,\\
  n_k^{+} &\sim Binomial(n_k, p_k) & k=1\ldots K
\end{align}
Since the prior and likelihood are conjugate, we can directly infer the posterior distribution of $p_k$ as $Beta(a + n_k^{+}, b+n_k^{-})$.
This leads to the analytical solution of the posterior distribution of $\pi$,
\begin{align}
  P(\pi | x_{1\ldots M}, S, y_S) = \sum_{k=1}^K w_k . Beta(a + n_k^{+}, b+n_k^{-}) \label{eqn:posterior_prevalence}
\end{align}
which is a mixture of Beta distributions.
Since the mean and variance of a Beta distribution is known in closed form, we have the mean and variance of the posterior of $\pi$ from which we can compute the UBP under a Normal assumption.
In fact, the estimate of UBP is simple enough that it has been deployed as a SQL query in a large-scale OSN.

\subsection{Hyperparameter selection in Bucketed-Beta-Binomial}\label{sec:BBB_reservations}

Two important considerations influence the choice of hyperparameters in our models.

\paragraph{Time varying prevalence}
One of the key internal applications of prevalence estimation in OSNs is to monitor sharp rises in prevalence.
Such rises usually signify the exploitation of a vulnerability, which might necessitate a deeper analysis and possible redesign of the ML classifier.
This requires rapid detection of vulnerabilities and poses a dilemma for the system design.
In general, we need a large sample size to get good estimates, but we only have a fixed number of samples per day. To get a large sample size, we need to aggregate data over a larger number of days which can effectively smoothen sharp rises in prevalence and make them harder to detect.

An analysis of fluctuations in prevalence over many years by domain experts (in a large OSN) has shown that each policy group has its own cadence.
However, $30$ days is considered the largest interval for which a smoothening of prevalence will still allow any major changes in prevalence to be noticeable.
This number implies that the typical sample size we can expect to get is roughly $30,000$ since most policy groups have around $1000$ labeled samples collected per day.

\paragraph{Exponential distribution of scores}
An important property of the scores is that the number of items with a given score drops off exponentially as the score increases.
This distribution of scores is consistent across all policy groups although each policy group has its own exponential rate and it appears to be very stable over the years.
We treat this score distribution as part of our prior knowledge.

\paragraph{K, a, and b}
The maximum sample size together with the distribution of scores has implications on our choice of the number of buckets $K$.
Ideally, we want $K$ to be large enough to capture the fluctuations in prevalence but not so many that we have buckets with no samples in them.
In practice, a value of $K$ between $5$ and $10$ works for all policy groups in ensuring that each bucket gets at least one sample.

For choosing $a$ and $b$ we fall back to the commonly used Laplace smoothing which prescribes a $Beta(a,b)$ distribution with $a=1$ and $b=1$.
An interpretation of Laplace smoothing is in terms of pseudo counts.
In this interpretation the $Beta(1,1)$ prior is seen as observing one extra positive and one negative sample.
Since we create $K$ Beta distributions, we attempt to preserve the total pseudo counts by setting $a=b=\frac{1}{K}$.

It is important to note that our hyperparameter choice is deployed for analyzing classifiers from all policy groups.
Policy violations can vary widely in prevalence and some of these can be fairly high prevalence although in this paper we are focusing on the low prevalence regimes.
Also ML classifiers can often produce non-monotonic calibration curves.
Practical considerations dictate that we pick the same hyperparameters for all policy groups.
Hence we can't make any asymmetric assumptions about $a$ and $b$ or that $a$ is increasing across buckets.

It is very tempting from an ML perspective to tune the hyperparameters based on MLE (maximum likelihood estimation)~\cite{fuchs1982maximum} or a similar principle.
However, such a tuning approach leads to a different effective model for each policy group for each day, and would be counter-productive in gaining trust in the predictions.
It is critical for the validation and adoption of our work to have a single model across all policies which is stable over multiple years.

\subsection{Bucketed-Gaussian Processes}
As discussed above, we ideally like to have a large enough value of $K$ to accurately model
fluctuations in the calibration curve.
In the bucketed approach, there is a limit to the maximum number of buckets because if we end up with too many buckets without observations then we fallback to the prior for those buckets, and are unable to
learn from the data. A natural extension of the bucketed model is to allow for the prevalence in nearby buckets to be a-priori correlated.
Modeling the correlation is robust to empty buckets since buckets borrow strength from each other.
We propose the Bucketed-Gaussian Processes (GP), which is an enhanced version of BBB that takes advantage of similarity between $f(s_{i})$ and $f(s_{j})$ where $s_{i}$ and $s_{j}$ are nearby scores.

In a GP \cite{Rasmussen06gaussianprocesses} the correlations of all the data points are modeled with each other. This approach would lead to a computational complexity of $O(N^3)$, which is infeasible for $N$ in the ballpark of $30,000$.
Instead, we propose to model the covariance between buckets which leads to a more reasonable computational cost of $O(K^3)$.
The covariance function depends on the distance between the mid-points of the buckets ($i$ and $j$ in Eq. \ref{eq:bucket_midpoint}).
This approach is similar to the inducing points method \cite{schulz18tutorial} which introduces small set of pseudo data points to summarize the actual data and
reduce the computation complexity. Mathematically,
\begin{align}
  & \mu \sim Normal(0, 1) \quad \mu \in \mathcal{R}, \\
  & n_k^+ \sim Binomial(n_k, p_k), \\
  & p_k = \Phi(q_k) \quad k=1,\ldots,K  \\
  & (q_1, \ldots, q_K) \sim \mathcal{N}(\mu, \Sigma) \quad \Sigma \in \mathcal{R}^{K*K}
  \intertext{Where}
  & \Sigma_{ij} = \alpha^2 \exp(\frac{-d_{ij}}{2 \rho^2}), \hspace{4mm} d_{ij} = (\frac{i - j}{K})^2 \label{eq:bucket_midpoint}
\end{align}
and $\Phi$ is the CDF of the standard normal distribution (although any function $\Phi(.): \mathbb{R} \rightarrow [0, 1]$ may be used), and $\alpha$, $\rho$ are scalar hyperparameters.

Note that because of using the covariance function,
Bucketed-GP \emph{is not sensitive to the choice of $K$}.
The constraint on $K$, the number of buckets, is imposed mainly by the computation limitation for calculating the covariance matrix inverse of size $K \times K$ as well as the need for $\frac{1}{K}$ to be much smaller than $\rho$ to accurately model the calibration curve.
In practice, for the studied OSN and for all policy groups, a value of $\rho=0.1$ captures the fluctuations in the calibration curve reasonably well, and so we fix $\rho$ to this value and set $K=100$.
For determining $\alpha$, we ran simulations from the GP prior with different values of $\alpha$ and picked the one that leads to a prevalence prior closest to uniform (or Beta(1,1)).
This value turns out to be $\alpha=1$.

With this model, we can use Markov Chain Monte Carlo (MCMC) methods to draw samples of $p_{1,\ldots,K}$ to estimate the posterior distribution of the prevalence.
For each sample, we compute our estimate of prevalence by,
\begin{align}
  \pi  = \sum_{k=1}^K w_k \cdot p_k
\end{align}
We use the empirical distribution of $\pi$ over all of the samples to get an estimate of $u_\pi$.

\section{Experiments}
\label{sec:results}
Our methods, exactly as described in this paper, have been deployed in a large scale OSN for nearly a year as of this writing and have shown consistently better results compared to the previous approach of using
bootstrapping.
In this section we show results on two low prevalence policy groups that were not available to us during the development of our work.
The results on these policies are representative of the improvement provided by our work in the low prevalence regime.
We omit results on the high prevalence regime where bootstrapping provides identical results to our work.

In the first experiment we set $K=5$ for BBB and $K=10$ in the second experiment.
For bootstrapping we draw $1000$ bootstrap samples for estimating prevalence.

\subsection{Simulation Procedure}
\label{sec:simulation}
Our validation approach is to simulate ground truth data from as realistic a distribution as possible.
We consider a couple of different policy groups and use the true distribution of scores $x_{1\ldots M}$ for each one.
In addition, we have obtained from domain experts for each policy, a representative calibration curve $f_{h}(.): [0, 1] \rightarrow [0, 1]$ that was hand-trained on historical data.
Together $\mathbf{x}$ and $f_h$ represent a ground truth prevalence,
\begin{equation*}\pi_{gt} = \frac{1}{M} \sum_{i=1}^M f_h(x_i).\end{equation*}

Our experiment workflow for each policy group consists of the following steps repeated for various values of $N$.
\begin{itemize}
  \item Repeat for $1000$ times the following,
\begin{itemize}
  \item Sample a dataset $S$ of size $N$ based on the weighted sampling strategy that is currently deployed in the OSN for the policy group.
  \item For each sampled item $i \in S$ sample a label, $y_i$ using $f_h$ as follows,
  \begin{equation*}
    y_i \sim Bernoulli(f_h(x_i)).
  \end{equation*}
  \item Using $\mathbf{x}$, $S$, and $y_S$ as inputs for each prevalence estimation methods (Bootstrapping, BBB and Bucketed-GP) compute ($l_\pi$, $u_\pi$), the $95\%$ confidence interval (CI) of $\pi$.
  \item Based on the CI calculate the following,
  \begin{itemize}
    \item \textbf{CI-Size} defined as $u_\pi - l_\pi$.
    \item \textbf{Wrong-CI} defined as $1$ if $\pi_{gt} \notin (l_\pi, u_\pi)$.
    \item \textbf{Wrong-Upper-Bound} defined as $1$ if $\pi_{gt} > u_\pi$.
  \end{itemize}
\end{itemize}
\item Compute average values of CI-Size, Wrong-CI, and Wrong-Upper-Bound for each of the estimation methods.
\end{itemize}

\subsection{Results on Semi-Real Data}
\begin{figure}[h!]
  \centering
    \includegraphics[width=0.85\columnwidth]{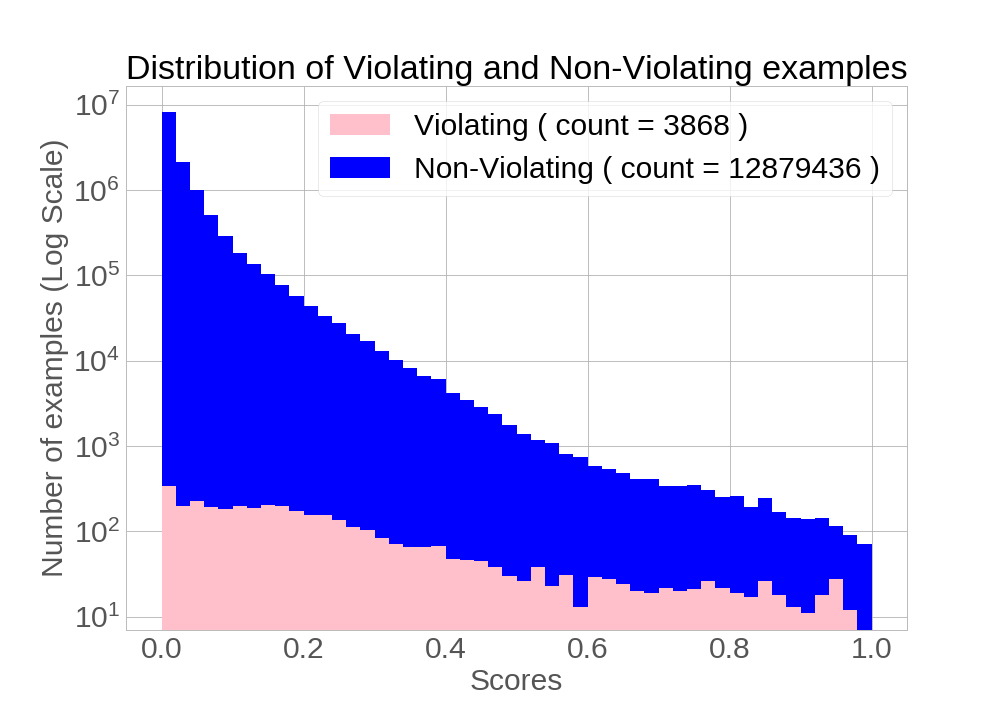}
    \includegraphics[width=0.85\columnwidth]{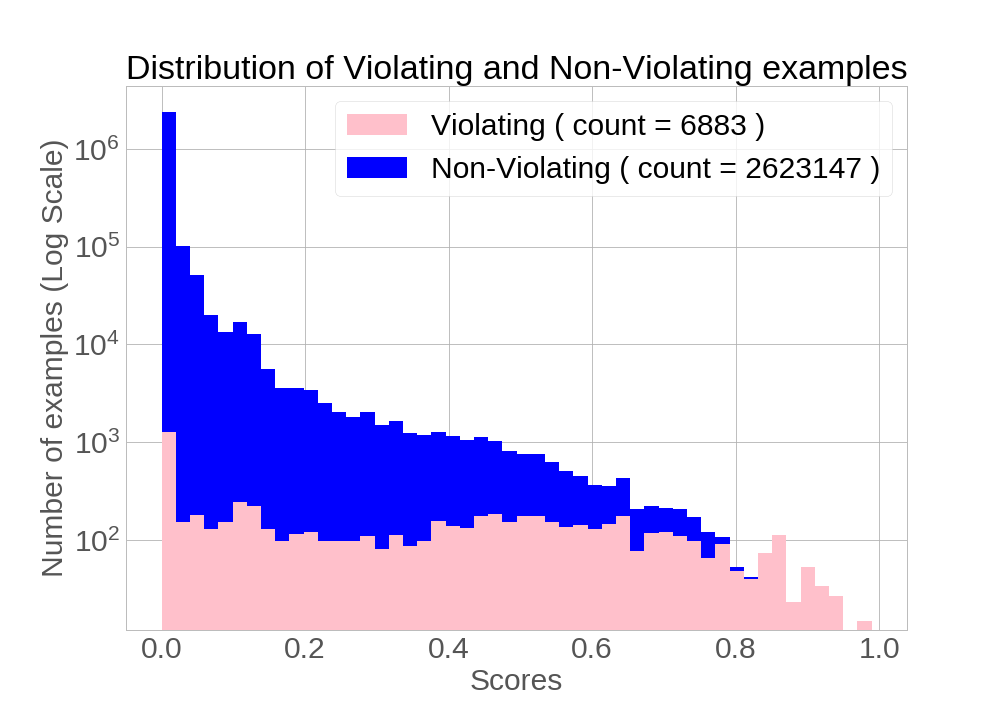}
  \caption{Distribution of scores in log scale for $\pi_{gt} = 3\times 10^{-4}$ (top) and $\pi_{gt}= 26\times 10^{-4}$ (bottom)}
  \label{fig:score_dist}
  \end{figure}

We analyzed the estimation procedures on two policy groups with prevalences as low as $3\times 10^{-4}$ and $26\times 10^{-4}$ with the distribution of labels as illustrated in Figure~\ref{fig:score_dist}.

\begin{figure}[h!]
  \centering
  \includegraphics[width=0.9\columnwidth]{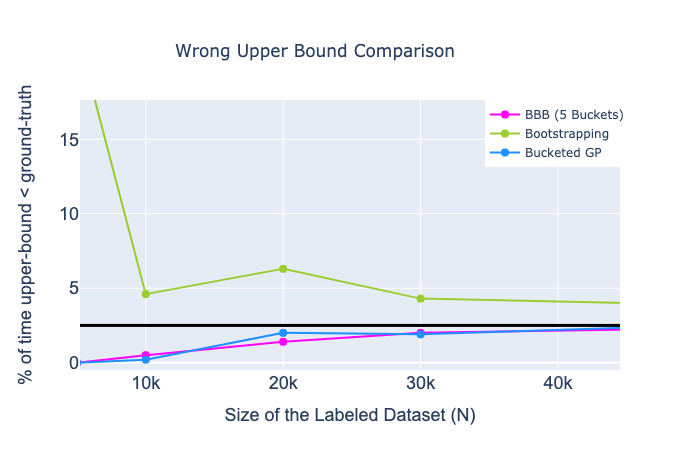}
  \includegraphics[width=0.9\columnwidth]{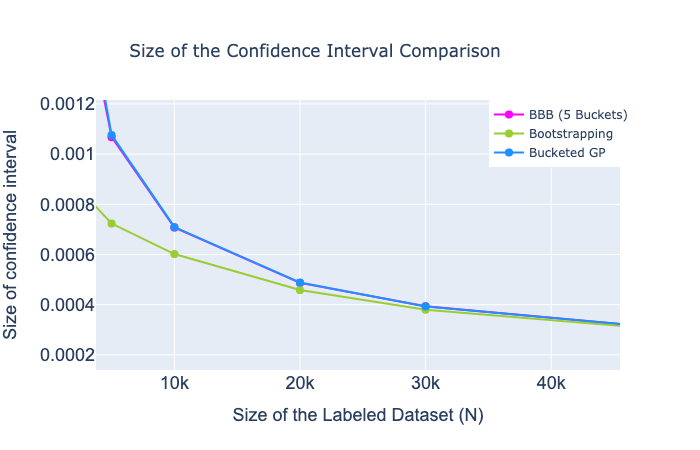}
  \includegraphics[width=0.9\columnwidth]{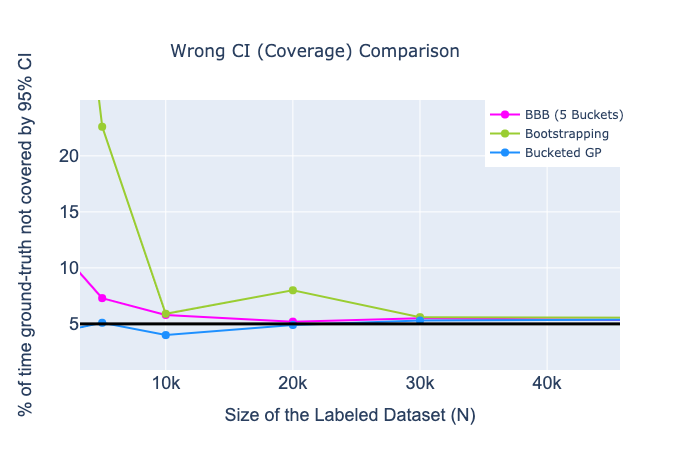}
  \caption{Prevalence estimation comparison for a policy with $\pi_{gt}=3\times 10^{-4}$; Wrong-Upper-Bound (top), CI-Size (middle) and Wrong-CI(bottom)}
  \label{fig:drug_sell_comparison}
  \end{figure}

Based on the discussion about the rolling window in Sec. ~\ref{sec:BBB_reservations}, UBP estimation is considered \textbf{practical} in our OSN only if it meets $2.5\%$ Wrong-Upper-Bound, $5\%$ Wrong-CI and reasonable CI-Size when
$N$ is roughly $30000$.
Figure ~\ref{fig:drug_sell_comparison} highlights the performance comparison of estimation methods as a function of $N$.

\textit{Wrong-Upper-Bound}: Top plot in Figure~\ref{fig:drug_sell_comparison} shows that Bootstrapping is typically above $2.5\%$ indicating that it consistently underestimates
UBP. For a dataset with $N \le 20000$ labels, where the odds of having zero violating labels is quite high, Bootstrapping hugely underestimates
UBP while Bucketed-GP and BBB tend to overestimate prevalence.
However, Bucketed-GP and BBB do well when we have more than $20000$
labels, hovering close to $2.5\%$ (horizontal black line in the plot).
Since our objective is reporting a reliable UBP, it is preferred to have a method that slightly overestimates prevalence than underestimates it.

\textit{Average CI-Size}: Middle plot in Figure~\ref{fig:drug_sell_comparison} shows that the CI-Size decreases as the number of labels increase, as expected.
This plot serves to act as a sanity check that our predictions don’t have an unreasonably large CI.
For example, for $N = 30000$, the average CI-Size is $4\times 10^{-4}$, which is on the
same order of magnitude as $\pi_{gt} = 3\times 10^{-4}$.


\textit{Wrong-CI}: Bottom plot in Figure~\ref{fig:drug_sell_comparison}  should be read alongside the other two plots because we could have a reasonably sized CI but it
is of little use if we are unable to cover the ground truth.
Evaluating the $95\%$ CI, we expect Wrong-CI to be around $5\%$ (horizontal black line
in the plot).
We see that after $20000$ samples, all methods start
to cover the ground truth, but with fewer samples, it is extremely hard for Bootstrapping to provide a good coverage.


Our overall results on this policy group show that we can in fact reduce the sample size to $20000$ for this policy and still maintain accuracy using our proposed methods while we couldn't have done that using the existing bootstrapping technique.
Reducing the sample size has important ramifications on our ability to quickly detect new trends in prevalence as discussed before, and these results confirm the positive impact of our work.

\begin{figure}[h!]
  \centering
\includegraphics[width=0.85\columnwidth]{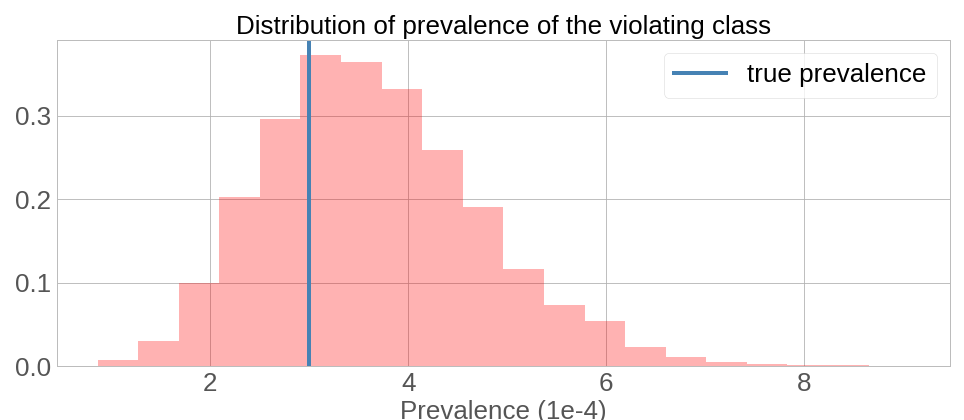}
\includegraphics[width=0.85\columnwidth]{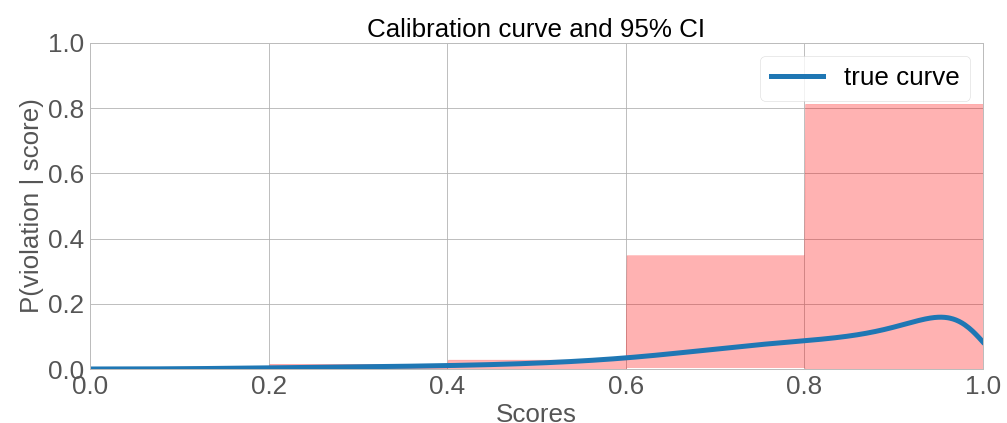}
\includegraphics[width=0.85\columnwidth]{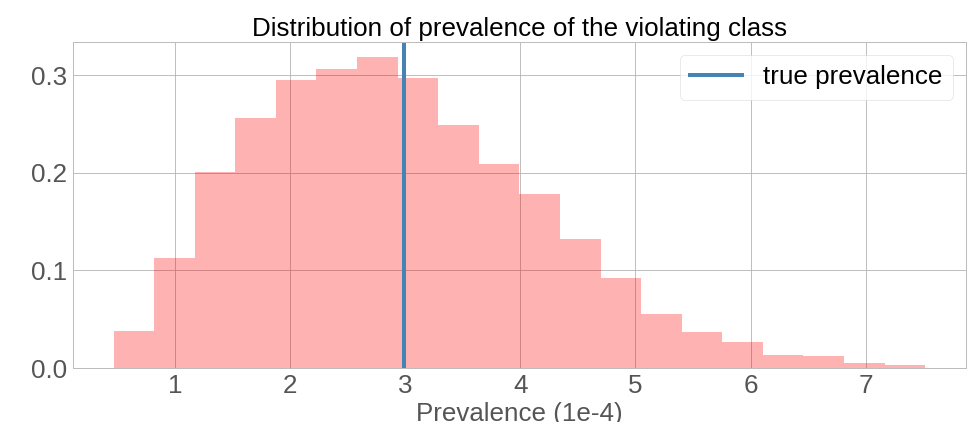}
\includegraphics[width=0.85\columnwidth]{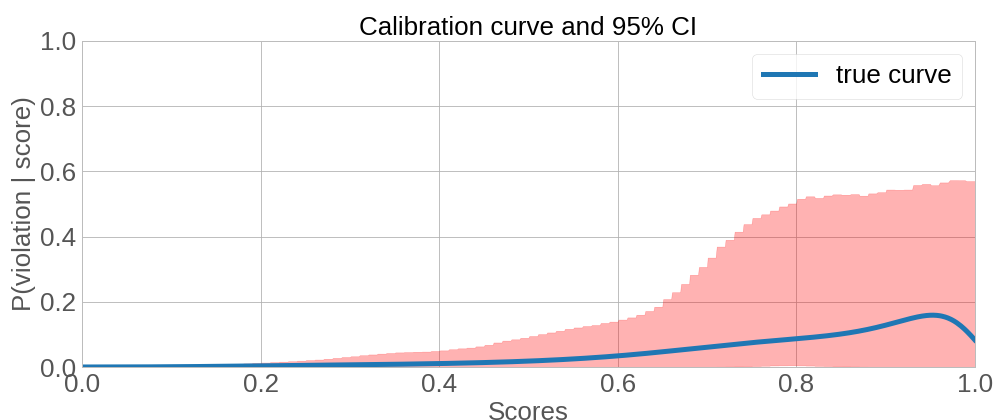}
\caption{Posterior prevalence distribution and calibration curves ($95\%$ CI) for BBB (top two) and Bucketed-GP (bottom two) for a policy with $\pi_{gt} = 3\times 10^{-4}$ and $N=30000$.}
\label{fig:drug_sell_uncertainty}
\end{figure}

In Figure~\ref{fig:drug_sell_uncertainty}, we show further validation of our Bayesian models by visualizing inferred posterior distributions of prevalence and the calibration curves.
It shows that both methods are able to cover the ground truth prevalence properly.
In the case of calibration curve, Bucketed-GP provides a higher resolution posterior than BBB, but both methods properly cover the ground truth calibration curve, $f_{h}(.)$. Note that a drop in the ground
truth calibration curve on the high end of the score range (Figure~\ref{fig:drug_sell_uncertainty}, second and fourth panel), is a typical artifact seen in many policy groups.
It is a consequence of the fact that a disproportionately high fraction of the truly violating content are identified and removed from the high end of the score range leading to a drop in prevalence for
the residual items in this range.

So far, in this paper, we have not discussed the sampling methodology and in general assumed that it is fixed per policy group.
However, it is worth pointing out briefly that the adoption of our work has opened up new avenues of research on the sampling front as well.
For example, the availability of the uncertainty bounds on the calibration curve and in particular the higher resolution bounds provided by Bucketed-GP allows for a dynamic adjustment of sampling weights on regions
of the score where the calibration has high uncertainty.
This dynamic sampling schema is a known technique in Bayesian Optimization ~\cite{frazier2018tutorial}, and is expected to lead to more precise estimates with fewer samples.

\begin{figure}[h!]
\centering
\includegraphics[width=0.9\columnwidth]{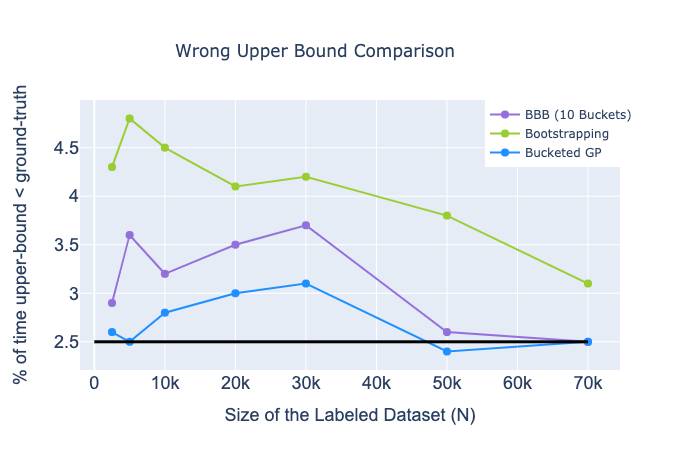}
\includegraphics[width=0.9\columnwidth]{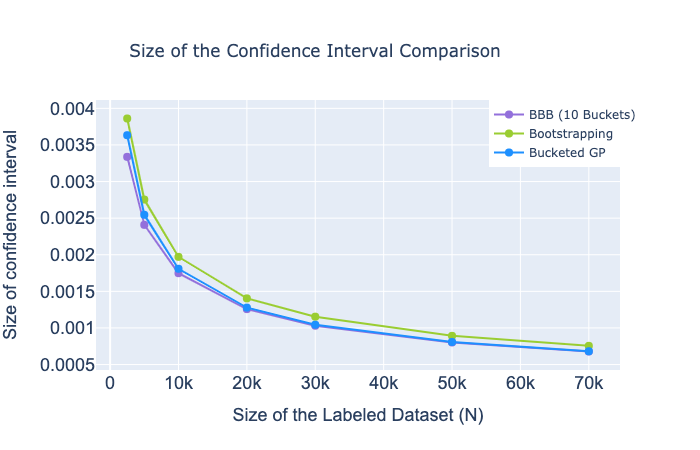}
\includegraphics[width=0.9\columnwidth]{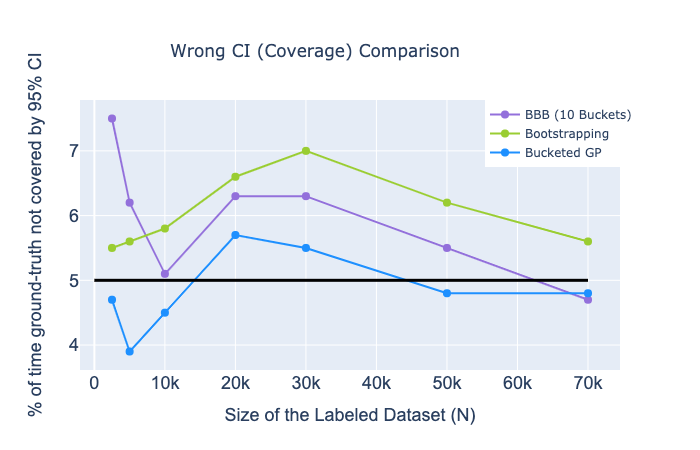}
\caption{Prevalence estimation comparison for a metric with $\pi_{gt} = 26\times 10^{-4}$; Wrong-Upper-Bound (top), CI-Size (middle) and Wrong-CI(bottom)}
\label{fig:hate_comparison}
\end{figure}

In Figure~\ref{fig:hate_comparison} we show results on a policy group that has an order of magnitude higher prevalence.
In this case, all methods are unable to get accurate UBP estimates for $N=30000$.
However, for $50000$ samples our methods achieve the desired accuracy while bootstrapping is still very far away from being accurate at this sample size, and needs at least $100000$ samples (not plotted in the figure) to work correctly.
It is important to note that a difference in this dataset from the previous one is that the true calibration curve has a much broader range of values.
Since our methods explicitly model this fluctuation in the curve they are able to give better predictions.
In particular, Bucketed-GP which has a finer resolution of the curve is able to give the best overall predictions.

\begin{figure}[h!]
  \centering
\includegraphics[width=0.85\columnwidth]{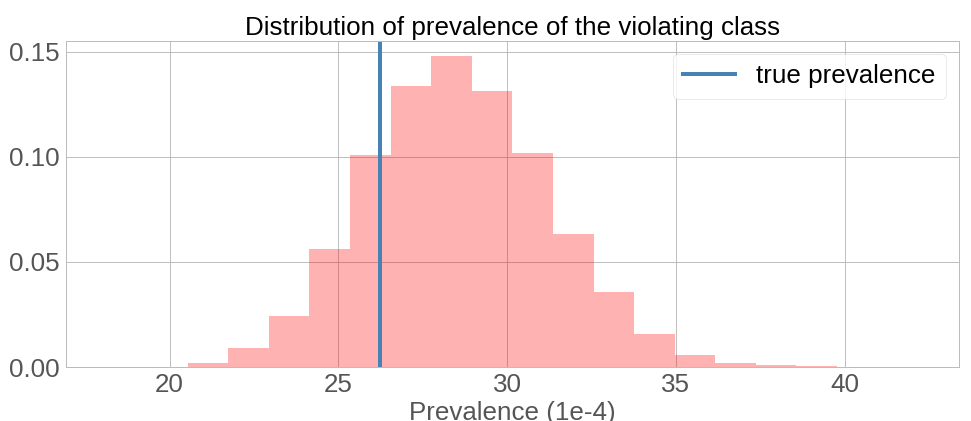}
\includegraphics[width=0.85\columnwidth]{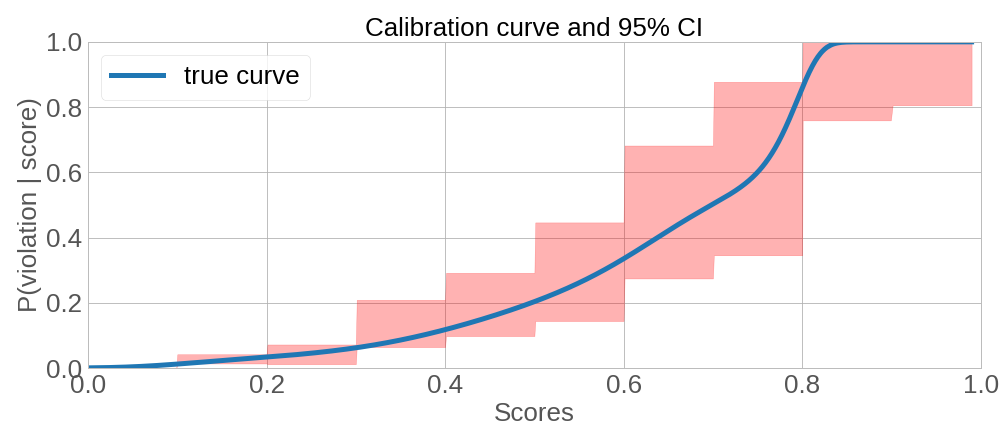}

\includegraphics[width=0.85\columnwidth]{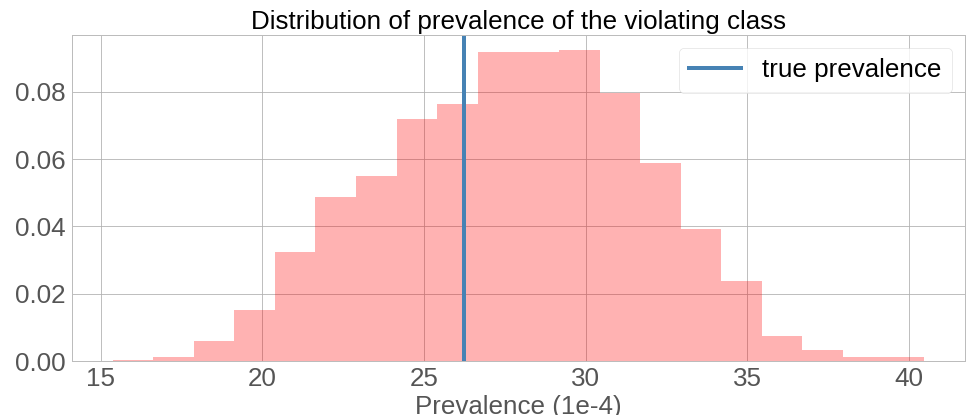}
\includegraphics[width=0.85\columnwidth]{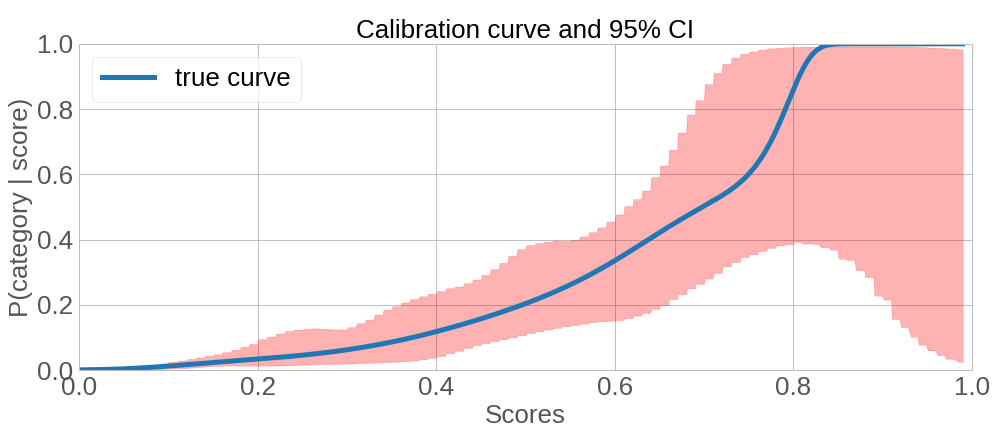}
\caption{Posterior prevalence distribution and calibration curves ($95\%$ CI) for BBB (top two) and Bucketed-GP (bottom two) for a policy with $\pi_{gt} = 26\times 10^{-4}$ and $N=30000$.}
\label{fig:uncertainty_hate}
\end{figure}

Finally, in Figure~\ref{fig:uncertainty_hate} we show the posterior distribution of prevalence and the calibration curves for this policy group.
Both methods produce reasonable posteriors, however, as before, Bucketed-GP has higher resolution than BBB.

\section{Conclusion}
\label{sec:conclusion}
Online Social Networks (OSNs) deal with a corpus of billions of content from which they need to identify and remove content that violates their community standards.
While this seems like a straight-forward ML classification task, the sheer scale of the problem makes it intractable to completely eliminate all violations.
Due to transparency and accountability needs, OSNs need to compute and report \emph{accurate conservative} estimates of the residual prevalence of these violations.
The methods that can be deployed for this estimation task are required to be generalizable across hundreds of policy groups and make minimal assumptions about the underlying ML classifiers. Under these strict constraints, we have proposed two Bayesian methods that don't need any hyperparameter tuning and give accurate results for a wide range in prevalence.
Indeed, for low prevalence policy groups, the results of this paper are far superior
to previously deployed bootstrapping methods, and need 2x fewer samples to achieve similar accuracy.
For high prevalence policy groups, our methods don't offer an improvement, but most importantly lead to no degradation either, and this has allowed our work to be a drop-in replacement for bootstrapping in a
large-scale OSN. In addition to providing more accurate prevalence estimates, our work has also provided uncertainty bounds of calibration curves which was not previously available.
This extra information has opened up avenues for future work such as dynamic sampling of content. One of our methods is simple enough to be deployed as a SQL query and be easily replicated.
The other method provides higher resolutions results, but is also easily implemented in commonly available software such as Stan ~\cite{stan2018}.
The simplicity of our models and the minimality of their assumptions suggest that these can be deployed on any OSN.
We hope that these improvements can lead to safer OSNs that can continue to serve humankind without hindrance for its communication needs.

\clearpage
\bibliography{bibfile}
\section*{Broader Societal Implications}
We have discussed the broad societal impact of OSNs in the main text, and also the potential negative effects of malicious actors seeking to exploit vulnerabilities in the OSN.
Our work is directly related to the accurate measurement and reporting of violations of the OSN's community standards policies.
As such we provide a gauge on the overall safety of an OSN.
Given the global scale of OSNs, an error in our methodology has the potential to have broad negative effects.

We seek to increase the robustness of our work by focusing on methodologies that make minimal assumptions and can be rationalized by general statistical techniques such as Laplace smoothing.
Instead of following the ML trend of building highly specialized models and continually improving upon them, we take the opposite position of building a model that works across \emph{all} policies and which remains stable over long periods of time.

We would encourage any OSN seeking to replicate our work to extensively validate these methods  before relying on them for internal accountability or regulatory reporting.
Along these lines, we would like to emphasize that the work described in this paper has been deployed for over a year in a very large OSN and has been independently validated by multiple ML teams during this period as well as for a $6$ month period prior to deployment.
We are sharing this work at this point only after it has proven useful in the integrity efforts within a large OSN.

\end{document}